\documentclass[conference]{IEEEtran}
\IEEEoverridecommandlockouts
\usepackage{cite}
\usepackage{bbm}
\usepackage{amsmath,amssymb,amsfonts,booktabs,float}
\usepackage[colorlinks,bookmarksopen,bookmarksnumbered,citecolor=green, linkcolor=red, urlcolor=green]{hyperref}

\usepackage{algorithmic}
\usepackage{graphicx}
\usepackage{textcomp}
\usepackage{xcolor}
\def\BibTeX{{\rm B\kern-.05em{\sc i\kern-.025em b}\kern-.08em
    T\kern-.1667em\lower.7ex\hbox{E}\kern-.125emX}}
\begin{document}

\title{Domain Adaptive Pose Estimation Via Multi-level Alignment\\

\thanks{This work was supported by the
	National Natural Science Fund of China under Grant
	No.62172222, the Postdoctoral Innovative Talent
	Support Program of China under Grant 2020M681609.
	
	* Corresponding author.}
}

\author{\IEEEauthorblockN{Yugan Chen}
\IEEEauthorblockA{\textit{School of computer science and engineering} \\
\textit{Nanjing University of Science and Technology}\\
Nanjing, China \\
chenyugan@njust.edu.cn}\\

\IEEEauthorblockN{Yalong Xu}
\IEEEauthorblockA{\textit{School of computer science and engineering} \\
	\textit{Nanjing University of Science and Technology}\\
	Nanjing, China \\
	xuyalong@njust.edu.cn}\\

\IEEEauthorblockN{Xiaoqi An}
\IEEEauthorblockA{\textit{School of computer science and engineering} \\
	\textit{Nanjing University of Science and Technology}\\
	Nanjing, China \\
	xiaoqi.an@njust.edu.cn}\\
\and
\IEEEauthorblockN{Lin Zhao*}
\IEEEauthorblockA{\textit{School of computer science and engineering} \\
\textit{Nanjing University of Science and Technology}\\
Nanjing, China \\
zhaolin@njust.edu.cn}\\

\IEEEauthorblockN{Honglei Zu}
\IEEEauthorblockA{\textit{School of computer science and engineering} \\
\textit{Nanjing University of Science and Technology}\\
Nanjing, China \\
zuhonglei@njust.edu.cn}\\

\IEEEauthorblockN{Guangyu Li}
\IEEEauthorblockA{\textit{School of computer science and engineering} \\
\textit{Nanjing University of Science and Technology}\\
Nanjing, China \\
guangyu.li2017@njust.edu.cn}
}
\maketitle

\begin{abstract}
Domain adaptive pose estimation aims to enable deep models trained on source domain (synthesized) datasets produce similar results on the target domain (real-world) datasets. The existing methods have made significant progress by conducting image-level or feature-level alignment. However, only aligning at a single level is not sufficient to fully bridge the domain gap and achieve excellent domain adaptive results. In this paper, we propose a multi-level domain adaptation approach, which aligns different domains at the image, feature, and pose levels. Specifically, we first utilize image style transfer to ensure that images from the source and target domains have a similar distribution. Subsequently, at the feature level, we employ adversarial training to make the features from the source and target domains preserve domain-invariant characteristics as much as possible. Finally, at the pose level, a self-supervised approach is utilized to enable the model to learn diverse knowledge, implicitly addressing the domain gap. Experimental results demonstrate  that significant improvement can be achieved by the proposed multi-level alignment method in pose estimation, which outperforms previous state-of-the-art in human pose by up to 2.4\% and animal pose estimation by up to 3.1\% for dogs and 1.4\% for sheep. The codes are available at \href{https://github.com/sunshinglight/DAML_PoseEstimation}{\textcolor{magenta}{the link.}}
\end{abstract}


\section{Introduction}
Recently, significant progress has been made in 2D pose estimation using deep learning methods \cite{sun2019deep,xiao2018simple,xu2022vitpose,yi2021transpose,zhao2021estimating}. However, these methods train high-performance models by using a large amount of labeled data, which is often time-consuming and expensive to obtain, especially with real-world datasets. To address the lack of the labels of the real-world datasets, domain adaptation (DA) \cite{saenko2010adapting} are being explored. By leveraging labeled virtual datasets, DA can transfer knowledge learned from virtual datasets (source domain) to unlabeled real-world datasets (target domain). Due to the development of computer technology and virtual reality techniques, labeled virtual data is much more cost-effective than real-world data. As a result, domain adaptive pose estimation has attracted a lot of attention.

While domain adaptation methods for classification tasks \cite{ganin2015unsupervised,hoffman2016fcns,long2015learning,zhang2019bridging} and semantic segmentation tasks \cite{hoyer2023mic,li2019bidirectional} have seen significant progress in recent years, there has been relatively little research applying these methods to pose estimation. Existing works, whether for human pose estimation or animal pose estimation, can be broadly classified into two categories. One focuses on narrowing the domain gap between domains by performing domain alignment on the image-level, while the other primarily focuses on the pose-level. UDAPE \cite{kim2022unified} proposes a style transfer technique that aims to align domains at the image-level. RegDA \cite{jiang2021regressive} employs adversarial training with two different regressors to correct pose errors. CC-SSL \cite{mu2020learning} and UDA-ANIMAL \cite{li2021synthetic} achieve pose-level alignment by designing different strategies to continuously update pseudo-labels. Although these methods all successfully improve the accuracy and performance of pose estimation, aligning domains at a single level fails to bridge domain gap comprehensively.

In order to comprehensively bridge the domain gap, in this work, we propose a multi-level alignment framework for DA pose estimation. The framework performs image, feature and pose levels alignment to effectively bridge the gap between different domains. First, at the image-level, the source image is transferred to the style of the target image, so that they can have similar data distributions. Second, at the feature-level, we utilize adversarial learning, which is achieved by incorporating a discriminator with gradient reverse layer, to align the distribution of features between two domains. This approach effectively makes model produces domain-invariant features. Lastly, at the pose-level, we utilize an information maximization self-supervised learning technique. This enables the model to learn meaningful and diverse pose representations and prevent it from biasing toward the source domain. We evaluate the proposed method on multiple datasets and demonstrate its effectiveness in bridging the domain gap. The main contributions of our work are as follows:
\begin{itemize}
	\item We propose a novel framework that leverages different alignment strategies at the image, feature, and pose levels to address the domain gap in cross-domain pose estimation.
	\item We conduct comprehensive experiments on both human and animal domain adaptive pose estimation benchmarks and achieve the state-of-the-art performance. For example, we achieve 84.4\% of accuracy on the task \textit{SURREAL} $\to$ \textit{LSP}, which is 2.4\% higher than the previous SOTA.
\end{itemize}

\section{RELATED WORK}

\subsection{Pose Estimation}

Pose estimation is a foundational visual task. Existing works can be mainly divided into two categories: CNN-based \cite{sun2019deep,xiao2018simple,zhao2021estimating} and transformer-based \cite{xu2022vitpose,yi2021transpose}. Simplebaseline \cite{xiao2018simple} introduces a simple and effective model based on ResNet \cite{he2016deep}.  HRNet \cite{sun2019deep} maintains high-resolution images throughout the entire training phase, achieving significant results. ViTPose \cite{xu2022vitpose} achieves excellent results by utilizing a regular ViT \cite{dosovitskiy2020image} structure as the backbone and combining it with a lightweight decoder. These methods have achieved near-human-level estimation results, but they require a large amount of  labeled data for training. Hence, in this work, we focus on addressing the domain gap issue in domain adaptive pose estimation, which makes it possible to train high-performance models solely using synthetic data.

\subsection{Domain Adaptive Pose Estimation}
Currently, there are two main categories of domain adaptation frameworks for pose estimation. The first category utilizes shared network structures, where the weights of the network are the same for both the source and target domains. RegDA \cite{jiang2021regressive} employs two independent regressors to narrow the domain gap adversarially. CC-SSL \cite{mu2020learning} utilizes consistency transformation and refinement of pseudo-labels. The second category involves non-shared network structures, typically employing a teacher-student paradigm to update weights. UDAPE \cite{kim2022unified} provides a good framework for DA pose estimation by style transfer. UDA-Animal \cite{li2021synthetic} combines pseudo-label updates with the mean-teacher \cite{tarvainen2017mean} framework. However, the aforementioned works only conduct alignment on a single level. Our work aims to achieve domain adaption through a multi-level approach.

\section{METHODOLOGY}
\begin{figure*}[t]
	\centering
	\setlength{\belowcaptionskip}{-10cm}
	\includegraphics[width=1\textwidth]{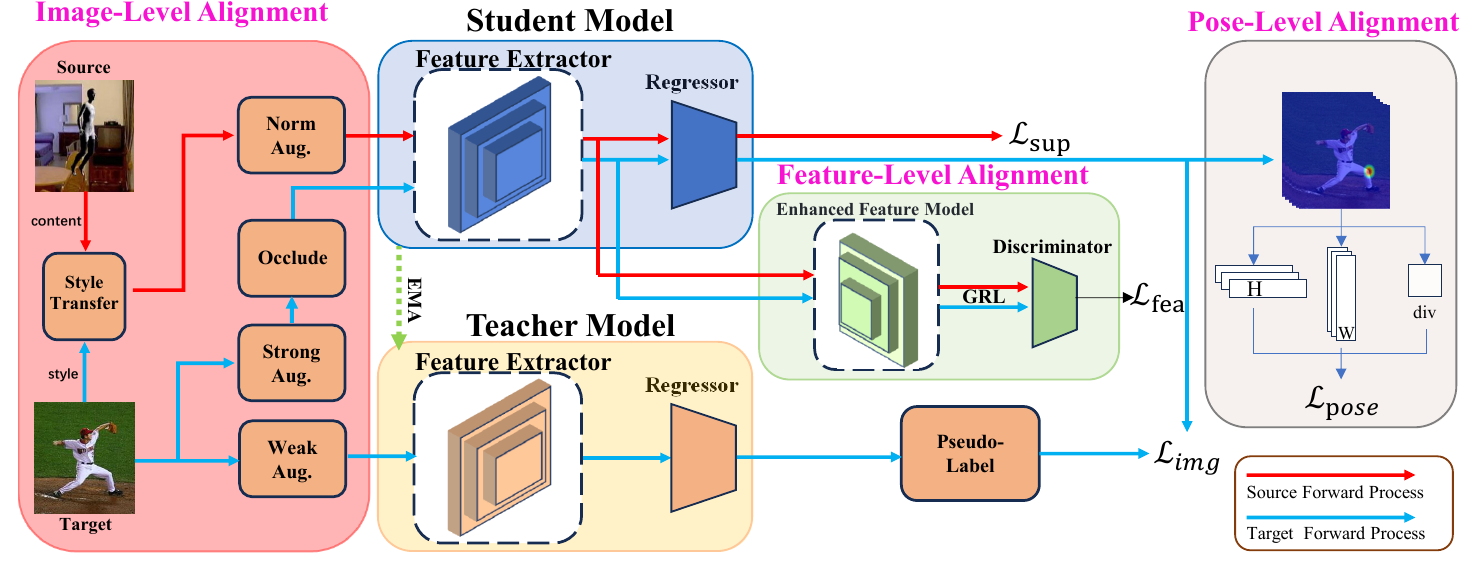}
	
	\caption{\textbf{Overview of our proposed method}. The framework consists of two branches: 1) Student model for cross-domain learning and 2) Teacher model that provides pseudo-labels for the Student model. We employ image-level alignment through style transfer during the input image processing, feature-level alignment through adversarial learning and pose-level alignment through self-supervised learning to update the student model's parameters, and use exponential moving averages (EMA) to update the teacher model. \textbf{GRL} refers to Gradient Reversal Layer, which is used to align the distributions of the two domains using a discriminator with gradient inversion layers.}
	\label{fig2:env}
\end{figure*}
\subsection{Preliminaries and Overview}
In the source domain, a labeled pose dataset $\mathcal{S}$ $\in$ $\left\{\left(x_s^i,y_s^i\right)\right\}_{i=1}^N$ includes $N$ samples, where $x_s^i\in \mathbb{R}^{C{\times}H{\times}W}$ represents a training image and $y_s^i{\in}\mathbb{R}^{K{\times}2}$ represents the corresponding keypoint coordinates. Here, $H$, $W$, $C$ and $K$ denote the height, width, the number of channels and the number of keypoints respectively. In the target domain, an unlabeled pose dataset $\mathcal{T}\in$$\left\{x_t^i\right\}_{i=1}^M$ includes $M$ samples. In most 2D pose estimation methods, a model is trained to output a set of heatmaps $\mathcal{H}{\in}\mathbb{R}^{K{\times}H^{\prime}{\times}W^{\prime}}$, where $H^{\prime}$
and $W^{\prime}$ are the height and width of the heatmap. The final keypoint coordinates are obtained by processing these heatmaps. Ground truth heatmaps are generated through a function (usually Gaussian) $G:\mathbb{R}^{K{\times}2}\to\mathbb{R}^{K{\times}H^{\prime}{\times}W^{\prime}}$. The entire training process is supervised using the MSE loss by comparing the predicted heatmaps with the ground truth.

The overview of the proposed DA framework is presented in Figure 1. Our model adopts the mean-teacher \cite{tarvainen2017mean} framework, which consists of a student model $f_s$ and a teacher model $f_t$. We train our model by using three alignment methods at the image, feature and pose levels respectively. To begin with, we pretrain a pose estimation model in the labeled source domain and let both the teacher and student share the same parameters from the pretrained model initially. During the image-level alignment, we make the source domain and the target domain have similar distribution by transferring the source image style to match the target image. Furthermore, we utilize the feature adversarial learning to align the distributions of features from two domains in student model, which will allow the student model to reduce domain gap. Finally, self-supervised learning is employed at the pose-level to reduce the student's bias towards the source domain and improve the teacher's ability to generate dependable pseudo-labels. 

\subsection{Image-Level Alignment}
Inspired by \cite{kim2022unified}, we utilize image style transfer to align the images from two domains. Firstly, a pre-trained VGG \cite{simonyan2014very} $f_v$ is used to extract corresponding features from the content image $x_s$ and style image $x_t$. Then, the generator $g$ in AdaIN \cite{huang2017arbitrary} is employed to generate the desired stylized output:
\begin{equation}
	\label{eq1}
	T(x_s,x_t,\eta) = g({\eta}t+(1-\eta)f_v(x_s)),
\end{equation}
where $t$ = AdaIN$\left(f_v(x_s),f_v(x_t)\right)$ and $\eta$ is a weight parameter controlling the ratio between the content image and style image. Through the style transfer method, we are able to transform the images from the source domain into $x_{s{\to}t} = T(x_s,x_t,\eta)$ and narrow the domain gap at the image level.

Moreover, to prevent the model from forgetting knowledge learned from the source domain during the learning process, we still need to supervise the student model using the labels of the source domain. After style transfer, we apply normal augmentation $\mathcal{A}_1$ to the style transferred images. Therefore, the supervised loss in the student model is:
\begin{equation}
	\label{eq3}
	\mathcal{L}_{sup} =\frac{1}{N}{\sum_{x_s \in \mathcal{S}}} \left\|{f_s(\mathcal{A}_1(x_{s{\to}t}))-\mathcal{H}_s} \right\|_2,
\end{equation}
where $\left\| \cdot \right\|_2$ refers to the MSE loss and $\mathcal{H}_s$ refers to the ground truth heatmap from the source domain.

To generate more reliable pseudo-labels, for the images from the target domain, we use different augmentation for the student model and teacher model. For the input of the student model, we first apply strong augmentation $\mathcal{A}_2$ to transform it, and then randomly occlude keypoints in the foreground to simulate various types of noise via an occluded operation: $\hat{x_t} = O({{A}_2}(x_t))$. For the input of the teacher model, we only use weak augmentation $\mathcal{A}_3$. Meanwhile, we set a random threshold $\tau_i$ to simulate the occurrence of random noise on the teacher side. Only when the activation value is greater than the set threshold $\tau_i$, do we perform consistent learning between the teacher and student models. Therefore, the loss function can be defined as follows:
\begin{equation}
	\label{eq4}
	\mathcal{L}_{img} ={\sum_{x_t \in \mathcal{T}}}{\sum_{k=1}^{K}}\mathbbm{1}(h_t^{\prime}{\geqslant}\tau_i) \left\|\tilde{\mathcal{A}_2}(h_t)-\tilde{\mathcal{A}_3}(h_t^{\prime})\right\|_2,
\end{equation}
where $\tilde{\mathcal{A}}$ denotes the reverse operation of ${\mathcal{A}}$, $h_t = f_s(\hat{x_t})$ denotes the outputs of the student model, $h_t^\prime = f_t({\mathcal{A}_2}(x_t))$ denotes the normalized output heatmap of the teacher model and $\mathbbm{1}(\cdot)$ denotes the function which can choose activation value. 
\subsection{Feature-Level Alignment}
Considering that pseudo-labels from the teacher model are influenced by the source domain, there is a risk of unintentionally biasing the student towards that domain. In order to make the model learn domain invariant feature representation, we apply adversarial learning. Through adversarial learning, the student model can leverage information from both domains to narrow the domain gap and align distributions before applying pseudo-label supervision.

The specific method involves using the student's feature encoder $F$ to obtain the initial features, and then using a feature enhancement model $f_e$ to remove noise from both domains and enhance more representative information. A feature discriminator $D$ is introduced to distinguish whether the features come from the source or the target domain. For each input sample, assume the probability of it belonging to the source domain as $D(f_e(F(\mathcal{X})))$, otherwise it is  $1 - D(f_e(F(\mathcal{X}))$. We use binary cross-entropy loss to update the discriminator,  and the domain label $d$ is set to 0 if the sample comes from the source domain, otherwise 1. The discriminator loss function can be represented as follows:
\begin{equation}
	\label{eq5}
	\mathcal{L}_{dis} = -d{\log}(D(f_e(F(\mathcal{X})))) - (1-d){\log}(1 - D(f_e(F(\mathcal{X})))),
\end{equation}
where $\mathcal{X}$ denotes the inputs from source or target domain after the augmentation.

Similar to GAN \cite{goodfellow2014generative}, we need to encourage our discriminator to distinguish the features from source or target domains as accurately as possible, while also ensuring that feature encoder can generate features that the discriminator cannot distinguish. Therefore, the final loss function can be defined as follows:
\begin{equation}
	\label{eq6}
	\mathcal{L}_{fea} = \max\limits_{F}\min\limits_{D}\mathcal{L}_{dis}.
\end{equation}

In this work, we utilize the method \cite{ganin2015unsupervised} to optimize the max-min problem by generating reverse gradients between the feature encoder and discriminator. 

\subsection{Pose-Level Alignment}
Due to the availability of labels only in the source domain
and the student model can access inputs from both the source
and target domains, the student model may exhibit a bias towards the source domain. To further narrow the gap between domains, we implicitly align domains at the pose level. Specifically, inspired by \cite{liang2020we}, an Information Maximization (IM) loss is used to achieve this goal. The IM loss aims to encourage the model to maximize the mutual information between the learned features from the input data, thereby making the model's output more deterministic at individual keypoints and more diverse on a global scale. This alignment does not involve explicit operations directly handling domain discrepancies, but it can improve the accuracy and robustness of the outputs of the student model. Therefore, it is referred to as implicit alignment. 

Following the concept introduced by \cite{peng2023source}, we decompose the heatmap into H-direction and W-direction vectors using projection vectors to mitigate the impact of keypoint sparsity. Additionally, considering that projection vectors may stack some irrelevant points together, we only apply the Information Maximization to points with confidence greater than a threshold $\tau_{p}$ to prevent the model from incorrectly shifting to non-keypoint areas. The self-supervised loss is defined as follows:
\begin{equation}
	\label{eq7}
	\mathcal{L}_{pose} = \mathbb{E}_{x^t_i\in\mathcal{T}}(\mathcal{L}_{enth} + \mathcal{L}_{entw} + \mathcal{L}_{div}),
\end{equation}
\begin{equation}
	\label{eq8}
	\left\{
	\begin{aligned} 
			\mathcal{L}_{enth}&=-{\sum_{k=1}^{K}}\mathbbm{1}(\varphi(h_t){\geqslant}\tau_{p})\phi(\varphi(h_t))\log\phi(\varphi(h_t)),  \\
			\mathcal{L}_{entw}&=-{\sum_{k=1}^{K}}\mathbbm{1}(\psi(h_t){\geqslant}\tau_{p})\phi(\psi(h_t))\log\phi(\psi(h_t)),\\
			\mathcal{L}_{div}&={\sum_{k=1}^{K}}\phi(h_t)\log\phi(h_t),
		\end{aligned}
		\right.
	\end{equation}
	where $\varphi(\cdot)$ denotes the operation to map $\cdot$ to a vector in direction of H,
	$\psi(\cdot)$ denotes the operation to map $\cdot$ to a vector in direction of W, and $\phi(\cdot)$ denotes the operation of softmax.
	\subsection{Training Details}
	In summary, the entire objective loss function can be defined as:
	\begin{equation}
		\label{eq9}
		\mathcal{L} = \mathcal{L}_{sup} + \alpha\mathcal{L}_{img} + \beta\mathcal{L}_{fea} + \gamma\mathcal{L}_{pose},
	\end{equation}
	where $\alpha$, $\beta$ and $\gamma$ are trade-off parameters.
	
	After conducting all the alignments, the student model can undergo normal gradient updates, while the parameters of the teacher are updated using the parameters of the student model through exponential moving average (EMA):
	\begin{equation}
		\label{eq2}
		{\theta}_t  \gets \mu {\theta}_{t} +(1-\mu){\theta}_s,
	\end{equation}
	where ${\theta}_t$ and ${\theta}_{s}$ denote the model parameters of the teacher
	and student, respectively. $\mu$ denotes the smoothing coefficient and the out default setting is 0.999.
	\section{Experiments}
	To validate the effectiveness of our approach on pose estimation tasks, we conduct experiments on benchmark datasets including human and animal pose estimation and compare our method with previous state-of-the-art models.
	
	\textbf{Datasets.} For human pose estimation, \textit{SURREAL} \cite{varol2017learning} and \textit{Leeds Sports Pose} (LSP) \cite{johnson2010clustered}  datasets are used. \textit{SURREAL} is a synthetic dataset used as the source domain, comprising a total of 6 million images. \textit{Leeds Sports Pose} is a real-world dataset containing 2k annotated images of athletes' poses. Our task is to perform domain adaptation from \textit{SURREAL} as the source domain to \textit{LSP} as the target domain.
	
	We utilized three datasets for animal pose estimation. The first one is \textit{SynAnimal} \cite{mu2020learning}, which is a synthetic dataset generated by rendering CAD models. It consists of five different animals: horse, tiger, sheep, hound and elephant. Each animal category contains 10k images. The second dataset is \textit{TigDog} \cite{del2015articulated}, a real-world dataset obtained by slicing frames from videos. It includes 30k images of horses and tigers. The third dataset is \textit{AnimalPose} \cite{cao2019cross}, which contains 6.1k real-world animal images of dogs, cats, cows, sheep, and horses. Our task is to perform domain adaptation from \textit{SynAnimal} as the source domain to \textit{TigDog} and \textit{AnimalPose} as the target domains.
	
	\textbf{Implementation Details.} The pose estimation network is based on SimpleBaseline \cite{xiao2018simple}. We use a pre-trained ResNet101 as the backbone network. The Adam \cite{kingma2014adam} optimizer is employed with an initial learning rate of 1e-4, which gradually decreases to 1e-5 at the 22500th iteration and finally reaches 1e-6 after 30000 iterations. As for hyperparameters, we set $\alpha$ = 1, $\beta$ = 0.1 and $\sigma$ = 0.3. After the training is completed, the teacher model is used for the final inference.

\subsection{Main Results}
In this section, the baselines, metrics and quantitative results on different tasks are shown.

\textbf{Baselines.} We compare our approach with the following baseline methods in domain adaptation for pose estimation: CC-SSL \cite{mu2020learning}, RegDA \cite{jiang2021regressive}, and UDAPE \cite{kim2022unified}. All baseline methods utilize ResNet-101 as the backbone.

\textbf{Metrics.} To evaluate our approach, the Percentage of Correct Keypoints (PCK) is used as the metric, which gives the percentage of correctly estimated keypoints. In Tables 1-5, we report PCK@0.05 that measures the percentage of accurate predictions, with a threshold of 5\% relative to the image size. For the human pose estimation task, we follow the same setting as other methods \cite{jiang2021regressive,kim2022unified} and focus on the following keypoints: Shoulder (Sld), Elbow (Elb), Wrist, Hip, Knee, and Ankle. As for the animal pose estimation tasks, we also follow the common settings as other methods \cite{li2021synthetic,kim2022unified}. In \textit{TigDog}, we select the Eye, Chin, Shoulder (Sld), Hip, Elbow (Elb), Knee, and Hoof. In \textit{AnimalPose}, the Eye, Hoof, Knee, and Elbow are selected.

\textbf{Results on Human Pose Estimation.} Table 1 presents the results of the human pose estimation task \textit{SURREAL} $\to$ \textit{LSP}. Our method achieves the best performance outperforming the previous state-of-the-art method UDAPE \cite{kim2022unified} by 2.4\%. Moreover, our method consistently surpasses the baseline methods in every keypoint for human pose estimation.

\textbf{Results on Animal Pose Estimation.} Tables 2-5 present the quantitative comparing results of animal pose estimation tasks: \textit{SynAnimal} $\to$ \textit{TigDog} and \textit{SynAnimal} $\to$ \textit{AnimalPose}. On \textit{TigDog}, our method achieves the best results on the tiger category (Table 2), surpassing UDAPE \cite{kim2022unified} by 0.5\%. On the horse category (Table 3), the performance of our method is on par with UDAPE and slightly lower than that of UDA-Animal. On \textit{AnimalPose}, our model achieves the best results on the dog and sheep categories, improving the previous best results by 3.1\% and 1.4\%, respectively.
\begin{table}[t]
	\begin{center}
		\caption{PCK@0.05 on task \textit{SURREAL} $\to$ \textit{LSP}} \label{tab:cap}	\vspace{-1.em}
		\resizebox{\linewidth}{!}{
			\begin{tabular}{lccccccc}
				\toprule
				method & Sld & Elb & Wrist & Hip & Knee & Ankle & All\\
				\midrule
				Source-only & 50.6 & 64.8 & 63.3 & 70.1 & 71.2 & 70.1 & 65.0\\
				\midrule
				
				CCSSL \cite{mu2020learning} (CVPR’20)  & 36.8 & 66.3 & 63.9 & 59.6 & 67.3 & 70.4 & 60.7\\
				UDA-Animal \cite{li2021synthetic}(CVPR’21) & 61.4 & 77.7 & 75.5 & 65.8 & 76.7 & 78.3 & 69.2\\ 
				RegDA \cite{jiang2021regressive} (CVPR’21)  & 62.7 & 76.7 & 71.1 & 81.0 & 80.3 & 75.3 & 74.6\\
				UDAPE \cite{kim2022unified} (ECCV’22)  & 69.2 & 84.9 & 83.3 & 85.5 & 84.7 & 84.4 & 82.0\\
				\midrule
				Ours & \textbf{78.2} & \textbf{86.6} & \textbf{83.7} & \textbf{87.1} & \textbf{85.2} & \textbf{85.5} & \textbf{84.4}\\
				\bottomrule
				
		\end{tabular}}
	\end{center}
	\vspace{-2.0em}
\end{table}
\begin{table}[H]
	\begin{center}
		\caption{PCK@0.05 on task \textit{SynAnimal} $\to$ \textit{TigDog (Tiger)}}
		\vspace{-1.em} \label{tab:cap}
		\resizebox{\linewidth}{!}{
			\begin{tabular}{lcccccccc}
				\toprule
				method &Eye & Chin & Sld & Hip & Elb & Knee & Hoof & All\\
				\midrule
				Source-only & 85.4 & 81.8 & 44.6 & 70.8 & 39.6 & 48.4 & 55.5 & 54.8\\
				\midrule
				CCSSL \cite{mu2020learning} (CVPR’20)  &94.3 &91.3 &49.5 &70.2 & \textbf{53.9} & 59.1 & 70.2 & 66.7\\
				RegDA \cite{jiang2021regressive} (CVPR’21)  & 93.3& 92.8& 50.3& 67.8& 50.2 &55.4 &60.7 &61.8\\
				UDA-Animal \cite{li2021synthetic}(CVPR’21) &98.4 &87.2& 49.4& \textbf{74.9}& 49.8& 62.0& 73.4& 67.7\\
				UDAPE \cite{kim2022unified} (ECCV’22)  &\textbf{98.5} &\textbf{96.9} &56.2 &63.7 &52.3 &62.8 &72.8 &67.9\\
				\midrule
				Ours & 98.0&95.4&\textbf{60.4}&64.1&52.1&\textbf{63.3}&\textbf{73.6}&\textbf{68.4}\\
				\bottomrule
		\end{tabular}}
	\end{center}
	\vspace{-2.0em}
\end{table}
\begin{table}[H]
	\vspace{-1.5em}
	\begin{center}
		\caption{PCK@0.05 on task \textit{SynAnimal} $\to$ \textit{TigDog (Horse)}}
		\vspace{-1.em}
		\label{tab:cap}
		\resizebox{\linewidth}{!}{
			\begin{tabular}{lcccccccc}
				\toprule
				method &Eye & Chin & Sld & Hip & Elb & Knee & Hoof & All\\
				\midrule
				Source-only & 82.0 & 90.0 & 59.2 & 79.5 & 65.8 & 66.9 & 57.7 & 67.4\\
				\midrule
				CCSSL \cite{mu2020learning} (CVPR’20)  &89.3 &92.6 &69.5 &78.1 &70 &73.1 &65 &73.1 \\
				RegDA \cite{jiang2021regressive} (CVPR’21)  &89.2 &92.3 &70.5 &77.5 &71.5 &72.7& 63.2& 73.2\\
				UDA-Animal \cite{li2021synthetic}(CVPR’21) & 86.9 &\textbf{93.7} &\textbf{76.4} &\textbf{81.9} &70.6 &\textbf{79.1} &\textbf{72.6} &\textbf{77.5}\\
				UDAPE \cite{kim2022unified} (ECCV’22)  & \textbf{91.3} & 92.5 &74.0 &74.2 &75.8 &77.0 &66.6 &76.4 \\
				\midrule
				Ours &82.3&92.7&72.6&69.2&\textbf{76.4}&77.8&68.5&76.3\\
				\bottomrule
		\end{tabular}}
	\end{center}
	\vspace{-1.5em}
\end{table}

\begin{table}[H]
	\vspace{-2em}
	\begin{center}
		\caption{PCK@0.05 on task \textit{SynAnimal} $\to$ \textit{AnimalPose (Dog)}}
		\vspace{-1.em} \label{tab:cap}
		\resizebox{\linewidth}{!}{
			\begin{tabular}{lccccc}
				\toprule
				method &Eye & Hoof & Knee & Elb & All\\
				\midrule
				Source-only & 38.2 &43.2& 25.7& 24.1& 32.0\\
				\midrule
				CCSSL \cite{mu2020learning} (CVPR’20)  &34.7 &37.4 &25.4 &19.6 &27.0 \\
				RegDA \cite{jiang2021regressive} (CVPR’21)  &46.8 &54.6& 32.9& 31.2& 40.6\\
				UDA-Animal \cite{li2021synthetic}(CVPR’21) & 26.2 &39.8 &31.6 &24.7 &31.1\\
				UDAPE \cite{kim2022unified}  (ECCV’22) & 56.1 &\textbf{59.2}& 38.9& 32.7& 45.4\\
				\midrule
				Ours & \textbf{70.6}&59.1&\textbf{40.2}&\textbf{35.2}&\textbf{48.5}\\
				\bottomrule
		\end{tabular}}
	\end{center}
	\vspace{-2.0em}
\end{table}
\begin{table}[h]
	\begin{center}
		\caption{PCK@0.05 on task \textit{SynAnimal} $\to$ \textit{AnimalPose (Sheep)}} \label{tab:cap}
		\vspace{-1.em}
		\resizebox{\linewidth}{!}{
			\begin{tabular}{lccccc}
				\toprule
				method &Eye & Hoof & Knee & Elb & All\\
				\midrule
				Source-only & 59.9& 60.7& 46.2& 31.0& 47.9\\
				\midrule
				CCSSL \cite{mu2020learning} (CVPR’20)   &44.3 &55.4 &43.5 &28.5 &42.8\\
				RegDA \cite{jiang2021regressive} (CVPR’21)  & 62.8& 68.5& 57.0& 42.4 & 56.9\\
				UDA-Animal \cite{li2021synthetic}(CVPR’21) & 48.2 &52.9 &49.9 &29.7 &44.9\\
				UDAPE \cite{kim2022unified}  (ECCV’22) &61.6& \textbf{77.4}& 57.7& 44.6& 60.2\\
				\midrule
				Ours &\textbf{66.8}&75.8&\textbf{61.6}&\textbf{44.8}&\textbf{61.6}\\
				\bottomrule
		\end{tabular}}
	\end{center}
	\vspace{-1.5em}
\end{table}

\subsection{Ablation Study}
We conduct ablation studies to investigate the effectiveness of each module proposed in our method.

\textbf{Effect of loss functions.} We evaluate the performance of each component of our framework through the task \textit{SURREAL} $\to$ \textit{LSP} in Table 6. We adopt the mean-teacher  method \cite{tarvainen2017mean} as our baseline. Building upon image-level alignment, we introduce additional loss functions for both feature-level alignment and pose-level alignment: adversarial loss and self-supervised loss respectively. Experimental results demonstrate that feature-level alignment significantly enhances the model's performance, resulting in a 0.7\% improvement. Pose-level alignment alone does not provide significant benefits, yielding a marginal improvement of 0.3\%. Nevertheless, when both pose-level and feature-level alignment are conducted, an additional 1.1\% improvement is observed. In summary, our proposed multi-level alignment strategy proves to be crucial in bridging the domain gap.
\begin{table}[!h]
	\vspace{-1em}
	\begin{center}
		\caption{ The effect of loss functions on task \textit{SURREAL} $\to$ \textit{LSP}} \label{tab:cap}
		\vspace{-1.em}
		\resizebox{\linewidth}{!}{
			\begin{tabular}{lccccccc}
				\toprule
				method & Sld & Elb & Wrist & Hip & Knee & Ankle & All\\
				\midrule			
				MT & 57.1 & 70.6 & 65.7 & 74.6 & 76.1 & 74.5 & 69.8\\
				
				$\mathcal{L}_{img}$ & 69.3 & 86.3 & 84.5 & 86.3 & 84.6 & 84.4 & 82.6\\
				$\mathcal{L}_{img}\&\mathcal{L}_{fea}$ & 72.8 & 86.0 & \textbf{84.9} & 85.9 & 85.0 & 85.0 & 83.3\\
				$\mathcal{L}_{img}\&\mathcal{L}_{pose}$  & 71.2 & 86.4 & 84.5 & 86.4 & 84.7 & 84.5 & 82.9 \\
				$\mathcal{L}_{img}\&\mathcal{L}_{fea}\&\mathcal{L}_{pose}$ & \textbf{78.2} & \textbf{86.6} & 83.7 & \textbf{87.1} & \textbf{85.2} & \textbf{85.5} & \textbf{84.4}\\
				\bottomrule
		\end{tabular}}
	\end{center}
	\vspace{-1.5em}
\end{table}
\begin{figure}[!h]
	\begin{minipage}{0.32\linewidth}
		\vspace{3pt}
		\centerline{\includegraphics[width=\textwidth]{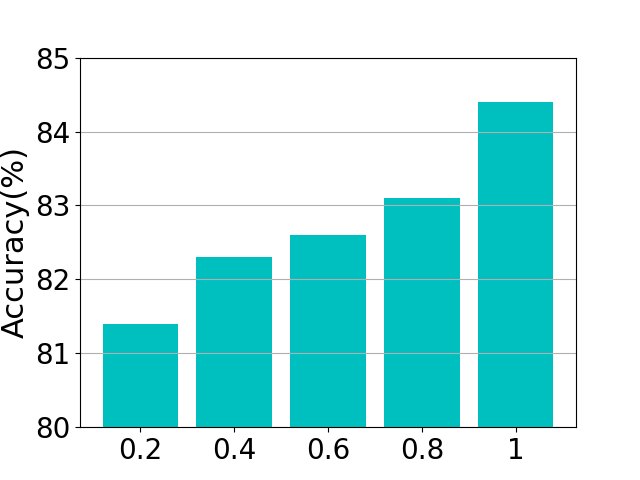}}
		\centerline{$\alpha$}
	\end{minipage}
	\begin{minipage}{0.32\linewidth}
		\vspace{3pt}
		\centerline{\includegraphics[width=\textwidth]{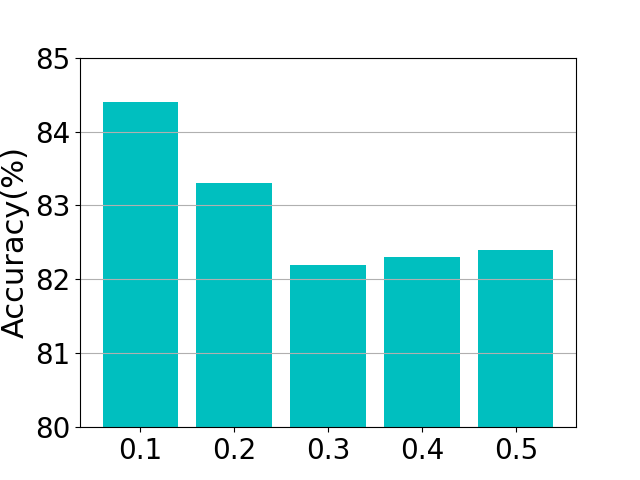}}
		
		\centerline{$\beta$}
	\end{minipage}
	\begin{minipage}{0.32\linewidth}
		\vspace{3pt}
		\centerline{\includegraphics[width=\textwidth]{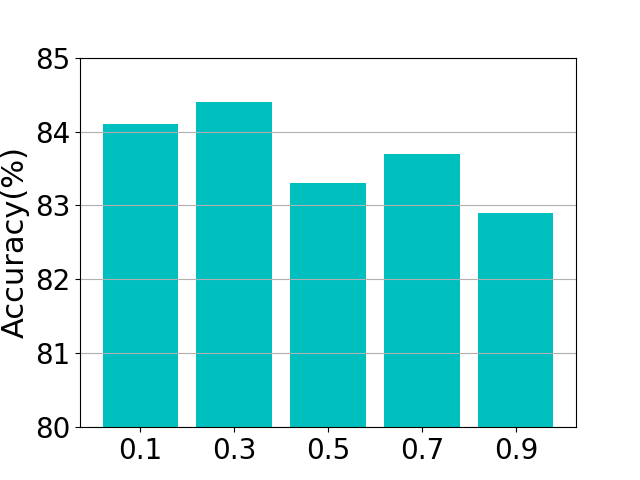}}
		
		\centerline{$\gamma$}
	\end{minipage}
	
	\caption{Analysis of the influences of parameters on \textit{SURREAL} $\to$ \textit{LSP}.}
	\label{fig4}
\end{figure}

\textbf{Analysis of parameters.} We illustrate the sensitivity of the parameters $\alpha$, $\beta$ and $\gamma$ in Equation 8, through the task \textit{SURREAL} $\to$ \textit{LSP}. The results are shown in Fig 2. Increasing 
$\alpha$ shows better performance and confirms the effectiveness of pre-alignment. However, increasing 
$\beta$ leads to a decline in performance and potentially causes overfitting. Lastly, the performance for $\gamma$ remains stable with little variation.
It can be observed that our model exhibits the best performance when $\alpha=1$, $\beta=0.1$ and $\gamma=0.3$.

\section{Conclusion}

In this paper, we propose a novel approach for unsupervised domain adaption on pose estimation. Based on the mean-teacher framework, we propose multi-level domain alignment strategy including image-level alignment through style transfer, feature-level alignment through adversarial learning and pose-level alignment through self-supervised learning method. The strategy effectively alleviates the domain gap and the source domain bias issues. The effectiveness and superiority of this approach have been verified in human and animal pose estimation domain adaption tasks.

\bibliographystyle{IEEEbib}
\bibliography{reference}

\end{document}